%% file: arxiv.tex
\documentclass[11pt]{article} %
\usepackage[utf8]{inputenc}
\usepackage{arxiv}
\usepackage{framed}
\usepackage{caption}
\usepackage{svg}
\usepackage{multirow}
\usepackage{multicol}
\usepackage{hyperref}
\usepackage{array}
\usepackage{longtable}
\usepackage{pythonhighlight}
\usepackage{booktabs}

\usepackage{geometry}
\usepackage{parskip}

\let\oldparagraph\paragraph
\renewcommand{\paragraph}[1]{\oldparagraph{#1.}}

\usepackage[T1]{fontenc}
\usepackage[scaled=.91]{helvet}
\usepackage{inconsolata}
\usepackage{algorithm}
\usepackage{algpseudocode}

\hypersetup{
    colorlinks=true,
    linkcolor=red,
    filecolor=magenta,      
    urlcolor=blue,
    pdfpagemode=FullScreen,
    }
    
\allowdisplaybreaks

\allowdisplaybreaks
\title{Interpretable Preferences via Multi-Objective Reward Modeling and Mixture-of-Experts}

\author{
Haoxiang Wang\thanks{Equal contribution. Correspondance to: Haoxiang Wang (\texttt{\href{mailto:hwang264@illinois.edu}{hwang264@illinois.edu}})}$^{\ast 1}$
~
Wei Xiong\footnotemark[1]$^{\ast 1}$
~
Tengyang Xie$^{2}$
~
Han Zhao$^{1}$
~
Tong Zhang$^{1}$
\\
\\
$^1$University of Illinois Urbana-Champaign $^2$University of Wisconsin–Madison
}
\date{}
\begin{document}
\maketitle

\begin{abstract}
Reinforcement learning from human feedback (RLHF) has emerged as the primary method for aligning large language models (LLMs) with human preferences. The RLHF process typically starts by training a reward model (RM) using human preference data. Conventional RMs are trained on pairwise responses to the same user request, with relative ratings indicating which response humans prefer. The trained RM serves as a proxy for human preferences. However, due to the black-box nature of RMs, their outputs lack interpretability, as humans cannot intuitively understand why an RM thinks a response is good or not. As RMs act as human preference proxies, it is desirable for them to be human-interpretable to ensure that their internal decision processes are consistent with human preferences and to prevent reward hacking in LLM alignment. To build RMs with interpretable preferences, we propose a two-stage approach: i) train an Absolute-Rating Multi-Objective Reward Model (ArmoRM) with multi-dimensional absolute-rating data, each dimension corresponding to a human-interpretable objective (e.g., honesty, verbosity, safety); ii) employ a Mixture-of-Experts (MoE) strategy with a gating network that automatically selects the most suitable reward objectives based on the context. We efficiently trained an ArmoRM with Llama-3 8B and a gating network consisting of a shallow MLP on top of the ArmoRM. Our trained model, \href{https://huggingface.co/RLHFlow/ArmoRM-Llama3-8B-v0.1}{\texttt{ArmoRM-Llama3-8B}}, obtains state-of-the-art performance on RewardBench, a benchmark evaluating RMs for language modeling. Notably, the performance of our model surpasses the LLM-as-a-judge method with GPT-4 judges by a margin, and approaches the performance of the much larger Nemotron-4 340B reward model. Our code and model are released at \url{https://github.com/RLHFlow/RLHF-Reward-Modeling}.
\end{abstract}

\section{Introduction}

In this paper, we explore the role of reward models (RMs) within the framework of Reinforcement Learning from Human Feedback (RLHF). RMs play a crucial role in aligning large language models (LLMs) as they provide a scalable way to integrate human preferences into the models' training process, guiding the optimization of their policies. To be more specific and provide more context, we first review the most standard and popular RLHF frameworks and the role of RMs in this framework. Arguably the dominant RLHF approach is a deep reinforcement learning (DRL)-based framework, as developed in key studies~\citep{christiano2017deep, ouyang2022training, bai2022training}. This framework operates in three stages: 1) Preference data collection; 2) Reward modeling based on the Bradley-Terry model \citep{bradley1952rank}; 3) Policy optimization using Proximal Policy Optimization (PPO) \citep{schulman2017proximal} and the reward model constructed in stage 2. This framework has achieved tremendous success in the post-training of ChatGPT \citep{ouyang2022training} and Claude \citep{bai2022training}. These ideas also extend to other approaches, such as rejection sampling fine-tuning \citep{dong2023raft, gulcehre2023reinforced} and iterative direct preference learning \citep{xiong2023iterative, guo2024direct, xie2024exploratory}. In these approaches, the intermediate policy is typically iteratively deployed to collect new responses, uses the reward model to label the responses, and fine-tunes the model on the newly collected preference data. In all of these RLHF frameworks, the capacity of the reward model is crucial as it directly affects the quality of the aligned LLMs.
\begin{figure*}[t!]
    \centering
    \includegraphics[width=1\linewidth]{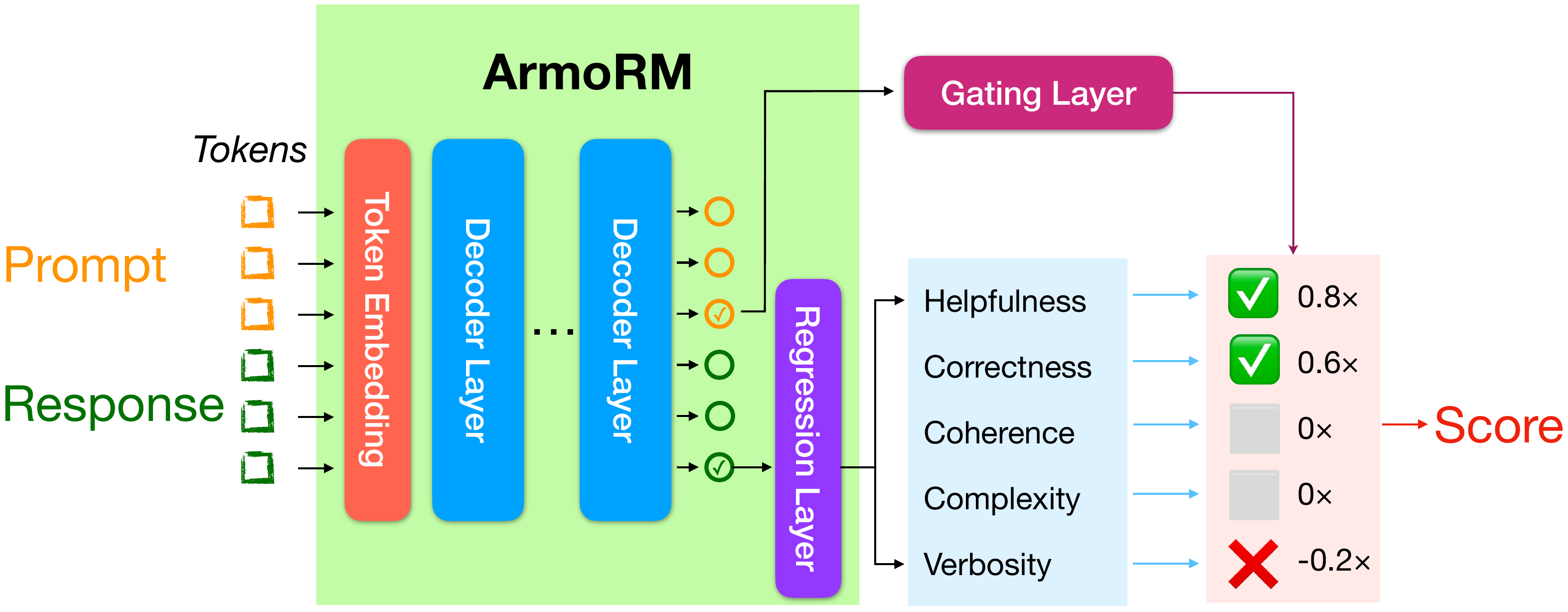}
    \caption{Architecture of our reward model. It consists of an LLM backbone, a regression layer for multi-objective reward modeling, and a gating layer that outputs coefficients to scalarize the reward objectives into a scalar score.}
    \label{fig:arch}
\end{figure*}

The most popular reward modeling approach is based on the maximum likelihood estimation (MLE) of the Bradley-Terry (BT) model \citep{bradley1952rank}. Despite its widespread use, the BT model is rather limited in the capacity of capturing the complicated human preference \citep{munos2023nash, swamy2024minimaximalist, ye2024theoretical}. In addition to the capacity issue, common RMs, like the BT model, are typically black-box models that output scores or preferences without providing human-interpretable explanations, making it subject to the widely observed phenomenon of reward hacking \citep{skalse2209defining,singhal2023long,chen2024odin}, where the aligned LLMs generate high-reward responses (rated by the RM) that do not align with actual human preferences~\citep{gao2023scaling, lin2023speciality, coste2023reward}. A notable example of this is the verbosity bias, where aligned LLMs produce longer-than-necessary responses because the RM favors length, regardless of quality \citep{singhal2023long,wang2024arithmetic, chen2024odin}.

In this work, we aim to enhance reward models by making them more \emph{interpretable} \citep{molnar2020interpretable} and \emph{steerable} \citep{wong2021leveraging}. Using the aforementioned verbosity bias as an example, suppose the RM's output is decomposable, meaning that it assigns a high score to a response due to two factors: 40\% for its helpfulness and 60\% for its length. In this case, we can see that the RM may suffer from the verbosity bias. Furthermore, if the RM is steerable, we could adjust its decision-making process to base its scoring 100\% on helpfulness. This would be regardless of response length, thus mitigating the verbosity bias. Enhancing the interpretability of RMs also allows humans to verify whether RMs have similar internal decision processes to humans when acting as proxies for human preferences. We believe that this human-AI interaction process could ensure that RMs are consistent with human values and preferences, making RM-aligned LLMs more reliable and robust.

At a high level, we propose a two-stage approach that first trains a multi-objective RM and then learns a gating layer that scalarizes reward objectives in a mixture-of-experts way. We then empirically validate its effectiveness by training such an RM with Llama-3 8B \citep{meta_llama3}, and obtain state-of-the-art performance on RewardBench, a benchmark to evaluate RMs.

\section{Related Works}\label{sec:related-works}
\subsection{RLHF Algorithms}
The PPO-based RLHF framework is first popularized in \citet{christiano2017deep} and further developed by \citet{bai2022training, ouyang2022training} to make ChatGPT and Claude, which leverages a reward model to provide feedback during the RLHF process. However, getting the PPO work is challenging in the context of LLMs \citep{choshen2019weaknesses, engstrom2020implementation}. Thus, much efforts have been made in proposing alternative approaches to the PPO, such as the REINFORCE algorithm variants \citep{li2023remax, shao2024deepseekmath}. Another popular approach is the reward-ranked fine-tuning algorithm (RAFT) \citep{dong2023raft, gulcehre2023reinforced} that was used in LLaMA2 \citep{touvron2023llama}, Llama-3 \citep{meta_llama3}, Qwen2 \citep{qwen2} and Apple Intelligence. To implement rejection sampling, we typically sample $n$ responses per prompt and use a reward model to rank them according to some criteria. Then, we fine-tune the model on the high-rank responses (e.g., the one with the highest reward value). This algorithm is a strong baseline, especially in reasoning tasks \citep{aksitov2023rest, havrilla2024teaching}. 
All approaches mentioned above leverage external reward models to provide supervision signals during the RLHF process.

There is also a line of works studying direct preference learning algorithms \citep{zhao2023slic, rafailov2023direct, azar2023general, tang2024generalized}, which bypasses traditional reward modeling to learn directly from preference datasets in a supervised manner (hence the name direct preference learning). Direct Preference Optimization (DPO) is the most representative one. However, the original DPO is an offline algorithm without further exploration of the environments. The subsequent studies demonstrate that the online iterative variants surpass the original DPO with large margins \citep{xiong2023iterative, liu2023statistical, xu2023some, rosset2024direct, guo2024direct, xie2024exploratory, zhang2024self, dong2024rlhf}. Specifically, we can iteratively deploy the intermediate policy to collect new responses and use the external reward model to label them, and further fine-tune the model on the newly collected preference data using the DPO objective. 

To summarize, \textit{all the existing popular RLHF algorithms require an external reward model to provide preference signals to achieve their best performance.}

\subsection{Reward modeling in RLHF}
Traditionally, reward models in RLHF have utilized the Bradley-Terry (BT) model for preference estimation \citep{bradley1952rank, ouyang2022training, bai2022training, wang2023enable, rafailov2023direct}. Despite its widespread use, the BT model's inability to handle complex, in-transitive preferences has been highlighted in recent studies \citep{munos2023nash, swamy2024minimaximalist, ye2024theoretical}. It is also argued that the DPO-aligned model can serve as a reward function to provide token-wise rewards \citep{rafailov2024r, zhong2024dpo}, which are still confined to the BT model. There are also works dropping the BT assumption and directly modeling the probability of response one being preferred over another one \citep{llm-blender-2023, zhao2023slic, liu2023statistical, dong2024rlhf}. These models are referred to as the pairwise preference model, as they take two responses as the input. Another line of work explores multi-objective reward models that attempt to capture the complicated human preferences more effectively \citep{touvron2023llama, wang2023helpsteer, wang2024arithmetic}. However, the integration of these multi-dimensional signals typically relies on naive methods such as linear combinations, indicating a need for more sophisticated techniques.

\section{Methodology}
\subsection{Multi-Objective Reward Modeling}
Most existing reward models for LLM alignment are trained with Bradley-Terry loss on pairwise data with annotated preferences \citep{bai2022training,touvron2023llama,ouyang2022training}, using the same approach as InstructGPT \citep{ouyang2022training}. The pairwise preference annotations are essentially binary labels, e.g., $\{0, 1\}$, indicating which response is preferred by the annotator. We call them relative ratings here. However, in some recent high-quality datasets, the relative ratings are converted from absolute ratings. For instance, UltraFeedback \citep{cui2023ultrafeedback} is curated with 5-objective absolute ratings: Overall Score, Instruction Following, Truthfulness, Honesty, and Helpfulness (each objective has 5 distinct ratings based on pre-defined rubrics). The dataset is further binarized into pairwise comparisons, using the Overall Score, or the average score of the remaining 4 objectives, for training reward models or DPO. The original ratings are fine-grained, as each objective has continuous integer rating scores (e.g., 1, 2, 3, 4, 5). However, the binarization process discards some fine-grained information. For example, a pair of examples with scores 1:5 is labeled in the same way as another pair with scores 2:3. It is not justified that discarding the fine-grained preference information is beneficial. Hence, we would like to include all fine-grained information for reward modeling.

As the training examples come with multi-objective ratings, the straightforward approach for learning with these ratings is multi-objective regression\footnote{This approach is also adopted in Directional Preference Alignment \citep{wang2024arithmetic} and HelpSteer \citep{wang2023helpsteer}.}. Here, we briefly introduce the training procedure. We consider each example to consist of a prompt $x$ (including contexts from previous conversation turns), response $y$, and a $k$-dimensional rating vector $r\in \mathbb{R}^{k}$, where each dimension corresponds to a reward objective such as helpfulness and truthfulness. Now, we take a pre-trained decoder-only LLM without the original output linear layer as the feature extractor $f_\theta$. We pass $x\oplus y$, the concatenation of $x$ and $y$, through the decoder layers and take the hidden state of the final decoder layer on the last token as a $d$-dimensional feature. Also, we attach a new linear regression layer $w\in \mathbb{R}^{d \times k}$ on top of $f_\theta$, which outputs a $k$-dimensional rating prediction. The model can be simply trained with regression loss:
\begin{align}
\min_{\theta, w} \mathbb{E}_{x,y,r\in D}\|w^\top f_\theta(x\oplus y) - r\|_2^2
\end{align}
\subsection{Mixture-of-Experts Scalarization of Reward Objectives}

An ArmoRM can predict multi-objective rewards for each response. However, the multi-dimensional outputs need to be reduced to a scalar for ranking or pairwise comparisons of test examples. A straightforward approach is to take a linear combination of multiple objectives~\citep{hu2024revisiting} as in the literature of multitask learning. However, using fixed combination coefficients is too rigid for complex application scenarios. For instance, for prompts that could easily trigger unsafe responses, the safety objective should be assigned a large coefficient, as we wish the reward model to rank unsafe responses lower than safe ones. For prompts for math problem assistance, the safety objective becomes less relevant, and the helpfulness-related objectives should be the primary focus.

With the insight mentioned above, we propose a MoE-style scalarization of reward objectives, conditioned on the prompt $x$. On the architecture level, we just need to follow the common MoE practice to add a gating layer, $g_\phi : \mathbb{R}^d \mapsto \{v\in \mathcal R^{k}\mid v_i\geq 0 ~\mathrm{and}~\sum v_i = 1 \}$, that outputs non-negative coefficients (summing up to 1) for the reward objectives based on the feature extracted from the prompt, $f_\theta(x) \in \mathbb{R}^d$, i.e., the hidden state on the last token of $x$. Notice that $f_\theta(x)$ is provided for free in the forward pass of $f_\theta(x\oplus y)$, making the pipeline inference-efficient.

The gating layer $g_\phi$ can simply be a shallow MLP (i.e., fully-connected network) that takes the prompt feature $f_\theta(x)$ and outputs a $k$-dimensional vector, followed by a softmax function to ensure the elements of the output vector are non-negative and summing up to 1.

However, most reward objectives are highly correlated with verbosity, which indicates a strong verbosity bias \citep{saito2023verbosity}. Using non-negative gating coefficients would make the final output inherit the bias. To resolve the issue, we adjust each reward objective, $r_i$, with a penalty using the verbosity reward objective,
\begin{align}\label{eq:reward-adjust}
    r_i' \leftarrow r_i - \lambda_i r_{\mathrm{verbose}}
\end{align}
where the penalty coefficient $\lambda_i$ is chosen such that for a proper correction metric (e.g., Pearson or Spearman correlation coefficient) and a reference data distribution $\mathcal D$,
\begin{align}\label{eq:corr}
    \mathrm{Corr}_{\mathcal D}(r_i', r_{\mathrm{verbose}}) = 0
\end{align}
The adjusted reward vector is denoted as $r'\in \mathbb{R}^k$. 

Finally, we multiply the gating coefficients to the multi-objective rewards, to obtain a scalar score $s$ for the response $y$ given prompt $x,$
\begin{align}
    R = g_\phi(f_\theta(x))^\top r'
\end{align}
To train the gating layer, we freeze the backbone and the regression layer, and only train the gating layer using the Bradley-Terry loss with an additional scaling variable, $\beta\in\mathbb{R}$,
\begin{align}
    \min_{\phi,\beta} \mathbb{E} \left[ -\log \frac{\exp(\beta R_{\mathrm{chosen}})}{\exp(\beta R_\mathrm{chosen})+\exp(\beta R_\mathrm{rejected})} \right]
\end{align}
where $R_{\mathrm{chosen}}$ and $R_{\mathrm{rejected}}$ are the preference scores for the chosen and rejected responses in each pairwise example, $(x, y_{\mathrm{chosen}}, y_{\mathrm{rejected}})$.

\section{Experiment}
\input{tables/reward-bench}
\paragraph{Implementation of ArmoRM} We use the Llama-3 8B \citep{meta_llama3} architecture and initialize the model backbone with parameters from a Bradley-Terry RM of Llama-3 8B trained by \citet{dong2024rlhf}. We append a linear layer to the backbone, and train it with regression loss while keeping the backbone frozen. The training involves 19 objectives (including helpfulness, correctness, verbosity, etc.) from 8 datasets, with details presented in Appendix \ref{supp:exp}.
\paragraph{Implementation of MoE} The gating layer is a ReLU MLP of 3 hidden layers with 1024 hidden units. For the correlation metric $\mathrm{Corr}$ in Eq. \eqref{eq:corr}, we adopt the Spearman correlation \citep{spearman1904}, and use UltraFeedback \citep{cui2023ultrafeedback} as the reference data distribution $\mathcal D$. The scaling variable $\beta$ is initialized with a value of 100, and the gating layer is trained with the LLM backbone kept frozen. The training is conducted on 10 pairwise preference datasets, with details in Appendix \ref{supp:exp}.

\paragraph{Software} Our training code is built with PyTorch \citep{pytorch}, HuggingFace's Transformers \citep{wolf2019huggingface} and Scikit-learn \citep{scikit-learn}.

\paragraph{Hardware} Training ArmoRM (the multi-objective reward modeling stage) only involves training the last linear layer (i.e., linear probing), so we save features extracted from the backbone locally and then conduct linear probing with Scikit-learn's linear regression solver on a CPU. For the MoE stage, we also save features locally, and then train the gating layer on a single NVIDIA A6000 GPU.

\paragraph{Hyperparameters} The gating layer is trained using the AdamW optimizer \citep{adamw} with a learning rate of 0.001 for 10,000 steps with a batch size of 1024. We also apply a cosine decay learning rate scheduler.

\paragraph{Evaluation Benchmark} RewardBench \citep{lambert2024rewardbench} is the first benchmark constructed to evaluate reward models for language modeling. It consists of a diverse set of tasks designed to assess the performance of reward models for LLM alignment, including four primary categories (Chat, Chat Hard, Safety, Reasoning) and a category of prior sets. Each category consists of multiple datasets with pairwise preference data, where each pair includes a chosen and a rejected text response. The overall score is computed as a weighted average over the five categories, where the four primary categories have weights 1.0 and the prior-sets category has weight 0.5.
\paragraph{Evaluation Results} Table \ref{tab:reward-bench} compares the performance of our approach (ArmoRM + MoE) against other reward models. Several key observations can be made from these results:
\begin{itemize}[leftmargin=*,align=left,noitemsep,nolistsep]
\item Our model significantly outperforms the Llama-3 8B Bradley-Terry RM, which provides the LLM backbone for our model. This demonstrates the effectiveness of our ArmoRM design and the MoE gating mechanism in improving the performance of reward models.
\item Our model also outperforms the LLM-as-a-judge approach \citep{zheng2023judging} with GPT-4 judges by a considerable margin, indicating that our model could be used as a cheaper replacement for GPT-4 in many annotation jobs.
\item Our model of 8B parameters has performance nearly on par with the Nemotron-4 340B RM \cite{wang2024helpsteer2}, a giant reward model of 340B parameters. This highlights the power and potential of our reward modeling approach.
\end{itemize}

\nocite{young2024yi}

\section{Conclusion}
In this work, we addressed the critical issue of interpretability in reward models for RLHF in the context of aligning LLMs with human preferences. We proposed a novel two-stage approach, consisting of an ArmoRM and a MoE strategy with a gating network. Our ArmoRM, trained with Llama-3 8B, achieved state-of-the-art performance on RewardBench, demonstrating the effectiveness of our reward modeling approach.

\newpage
\bibliographystyle{plainnat}
\bibliography{reference}

\newpage
\appendix
\section{Experimental Details}\label{supp:exp}

\paragraph{Licenses} The model we use and fine-tune follows the Meta Llama3 license. All the datasets we use are open-sourced and can be used for research purposes (some could be used for commercial purposes, such as HelpSteer \citep{wang2023helpsteer}). 

\paragraph{Personally Identifying Info or Offensive Content}
For all datasets used in this work, according to their data curation process descriptions, they do not contain any information that names or uniquely identifies individual people, except for some examples that contain celebrity names. However, BeaverTails \citep{ji2023beavertails}, PKU-RLHF \citep{ji2023beavertails}, and HH-RLHF \citep{bai2022training,ganguli2022red} contain offensive content, which is deliberately selected to build human preference datasets that aim to teach LLMs which responses are safe to generate.

\paragraph{Multi-Objective Training Datasets}
In the stage of multi-objective reward modeling, we use training datasets with corresponding reward objectives detailed below.
\begin{itemize}[leftmargin=*,align=left,noitemsep,nolistsep]
    \item \textbf{HelpSteer} \citep{wang2023helpsteer} (35k data): 
        \begin{itemize}[leftmargin=*,align=left,noitemsep,nolistsep]
            \item helpsteer-helpfulness
            \item helpsteer-correctness
            \item helpsteer-coherence
            \item helpsteer-complexity
            \item helpsteer-verbosity (This is the verbosity objective we use in Eq. \eqref{eq:reward-adjust} and \eqref{eq:corr})
        \end{itemize}
    \item \textbf{UltraFeedback} \citep{cui2023ultrafeedback} (240k data): 
        \begin{itemize}[leftmargin=*,align=left,noitemsep,nolistsep]
            \item ultrafeedback-overall-score
            \item ultrafeedback-instruction-following
            \item ultrafeedback-truthfulness
            \item ultrafeedback-honesty
            \item ultrafeedback-helpfulness
        \end{itemize}
    \item \textbf{BeaverTails-30k} \citep{ji2023beavertails} (30k data): 
        \begin{itemize}[leftmargin=*,align=left,noitemsep,nolistsep]
        \item beavertails-is-safe
        \end{itemize}
    \item \textbf{CodeUltraFeedback} \citep{codeultrafeedback} (50k data): 
        \begin{itemize}[leftmargin=*,align=left,noitemsep,nolistsep]
            \item code-complexity
            \item code-style
            \item code-explanation
            \item code-instruction-following
            \item code-readability
        \end{itemize}
    \item \textbf{Prometheus} \citep{prometheus} (200k data): 
        \begin{itemize}[leftmargin=*,align=left,noitemsep,nolistsep]
            \item prometheus-score
        \end{itemize}
    \item \textbf{Argilla-Capybara}\footnote{\url{https://hf.co/datasets/argilla/Capybara-Preferences-Filtered}} \citep{daniele2023amplify-instruct} (15k data): 
        \begin{itemize}[leftmargin=*,align=left,noitemsep,nolistsep]
            \item argilla-overall-quality
        \end{itemize}
    \item \textbf{Argilla-OpenOrca}\footnote{\url{https://hf.co/datasets/argilla/distilabel-intel-orca-dpo-pairs}} (13k data): 
    \begin{itemize}[leftmargin=*,align=left,noitemsep,nolistsep]
            \item argilla-judge-lm
        \end{itemize}
    \item \textbf{Argilla-Math-Preference}\footnote{\url{https://hf.co/datasets/argilla/distilabel-math-preference-dpo}} (2.4k data): This dataset shares the objective ultrafeedback-instruction-following with UltraFeedback
\end{itemize}
\paragraph{Multi-Objective Data Pre-processing} When merging multiple datasets with absolute ratings (e.g., UltraFeedback and HelpSteer), we observe some issues with the data. Here, we present the issues and our approach to tackle them:
\begin{itemize}[leftmargin=*,align=left,noitemsep,nolistsep]
    \item \textbf{Different Rating Scales}: Different datasets may have different scales for the ratings. For instance, HelpSteer has a rating scale of 0-4, while UltraFeedback’s is 1-10. We linearly transform all ratings to make them between 0 and 1. For BeaverTails with True/False ratings (indicating safe or unsafe), we treat True as 1 and False as 0.
    \item \textbf{Similar Objectives}: There are some very similar objectives from different datasets. For example, the Helpfulness objective appears in both HelpSteer and UltraFeedback, and the Correctness objective of HelpSteer is quite similar to the Truthfulness of UltraFeedback. After carefully examining the datasets, we decided to treat similar objectives as separate objectives, as they are rated by different judges following different rubrics. For instance, data from HelpSteer are rated by 200 U.S.-based human annotators following customized rubrics, and UltraFeedback data are labeled with GPT-4 following another set of rubrics.
    \item \textbf{Missing Labels of the Merged Dataset}: When merging multiple datasets, each example of the merged dataset only has a subset of ratings; for example, each example from HelpSteer only has 5 ratings originating from the HelpSteer dataset, and it does not have ratings for other objectives (e.g., the objectives from UltraFeedback or BeaverTails). Hence, when optimizing the regression loss, we simply ignore the missing rating dimensions of each example and only compute the loss on the remaining dimensions.
\end{itemize}

\paragraph{Training Data of MoE} In the stage of the gating layer, we use the following preference datasets: 
\begin{itemize}[leftmargin=*,align=left,noitemsep,nolistsep]
    \item \textbf{HelpSteer} \citep{wang2023helpsteer} (37k pairs)
    \item \textbf{UltraFeedback} \citep{cui2023ultrafeedback} (340k pairs)
    \item \textbf{SHP} \citep{SHP} (93k pairs)
    \item \textbf{HH-RLHF} \citep{bai2022training,ganguli2022red} (157k pairs) 
    \item \textbf{PKU-SafeRLHF-30K} \citep{ji2023beavertails}
    \item \textbf{Argilla-Capybara} (15k pairs)
    \item \textbf{Argilla-Math-Preferences} (2.4k pairs)
    \item \textbf{CodeUltraFeedback} \citep{codeultrafeedback} (50k pairs)
    \item \textbf{PRM-Phase-2} \citep{lightman2023let} (80k pairs)
    \item \textbf{Prometheus2-Preference-Collection} \citep{kim2024prometheus} (200k pairs)
\end{itemize}

\paragraph{Preference Data Pre-processing} For datasets that are not binarized into response pairs (e.g., HelpSteer, UltraFeedback, SHP), we take the binarized versions pre-processed in \citet{dong2024rlhf}.

\newpage

\end{document}

%% file: tables/reward-bench.tex
\begin{table*}[t]
\caption{Performance comparison on RewardBench. The benchmark consists of four primary categories (weight 1.0) and one category of prior sets (weight 0.5). The weighted average accuracy is computed as the overall score.}\label{tab:reward-bench}
    \resizebox{\linewidth}{!}{
\begin{tabular}{ll|c|ccccc}
\toprule
\multicolumn{1}{c}{\textbf{Method}} &
  \multicolumn{1}{c}{\textbf{Base Model}} &
  \multicolumn{1}{c}{\textbf{Score}} &
  \multicolumn{1}{c}{\textbf{Chat}} &
  \multicolumn{1}{c}{\textbf{Chat Hard}} &
  \multicolumn{1}{c}{\textbf{Safety}} &
  \multicolumn{1}{c}{\textbf{Reasoning}} &
  \multicolumn{1}{c}{\textbf{Prior Sets} {\small (0.5 weight)}} \\
  \midrule
HelpSteer2 RM & Nemotron-4 340B & \textbf{89.3} &  95.8  & \textbf{87.1} & 91.5 & 93.7 & 67.4\\
\textbf{ArmoRM + MoE}     & Llama-3 8B  & \textbf{89.0}   & 96.9 & 76.8 & \textbf{92.2} & \textbf{97.3} & 74.3 \\
HelpSteer2 RM & Llama-3 70B & 86.3 &  91.3 & 80.3 & 92.8 & 90.7 & 66.5  \\
Preference Model & Llama-3 8B  & 85.7 & 98.3 & 65.8 & 89.7 & 94.7 & 74.6 \\
LLM-as-a-judge   & GPT-4 Turbo & 84.2 & 95.3 & 74.3 & 87.2 & 86.9 & 70.9 \\
LLM-as-a-judge   & GPT-4o  & 83.3 & 96.6 & 70.4 & 86.7 & 84.9 & 72.6 \\
Bradley-Terry    & Llama-3 8B  & 83.6 & \textbf{99.4} & 65.1 & 87.8 & 86.4 & \textbf{74.9} \\
Bradley-Terry    & Yi-34B      & 81.4 & 96.9 & 57.2 & 88.2 & 88.5 & 71.4\\
\bottomrule
\end{tabular}
}
\vspace{-0.8em}
\end{table*}